%% file: CAMERA_READY_CoNLL_May18.tex
\documentclass[11pt]{article}

\usepackage[final]{acl}

\usepackage{amsmath}
\usepackage{tabularx}
\usepackage{microtype}
\usepackage{booktabs}
\usepackage{enumitem}
\usepackage{multirow}
\usepackage[edges]{forest}
\usepackage{tikz-dependency}
\usepackage[english,bidi=default]{babel} 
\babelfont{rm}{TeXGyreTermesX} 
\usepackage{fontawesome5}

\usepackage{caption}
\title{What Exactly do Children Receive in Language Acquisition? A Case Study on CHILDES with Automated Detection of Filler-Gap Dependencies}

\author{
    Zhenghao Herbert Zhou$^{1}$ \hspace{1em} William Dai$^{2}$ \hspace{1em} Maya Viswanathan$^{2}$ \\
    {\bf Simon Charlow$^{1,3}$ \hspace{1em} R. Thomas McCoy$^{1,3}$ \hspace{1em} Robert Frank}$^{1,3}$\\
    $^{1}$Department of Linguistics, Yale University\\ 
    $^{2}$Department of Computer Science, Yale University\\
    $^{3}$Wu Tsai Institute, Yale University\\
    \texttt{\{herbert.zhou, william.dai, maya.viswanathan,}\\
    \texttt{simon.charlow, tom.mccoy, robert.frank\}@yale.edu}
}

\begin{document}

\maketitle
\begin{abstract}
  Children’s acquisition of filler-gap dependencies has been argued by some to depend on innate grammatical knowledge, while others suggest that the distributional evidence available in child-directed speech suffices. Unfortunately, the relevant input is difficult to quantify at scale with fine granularity, making this question difficult to resolve. 
  We present a system that identifies three core filler-gap constructions in spoken English corpora --- matrix wh-questions, embedded wh-questions, and relative clauses --- and further identifies the extraction site (i.e., subject vs.\ object vs.\ adjunct). 
  Our approach combines constituency and dependency parsing, leveraging their complementary strengths for construction classification and extraction site identification. 
  We validate the system on human-annotated data and find that it scores well across most categories.
  Applying the system to 57 English CHILDES corpora, we are able to characterize children’s filler-gap input and their filler-gap production trajectories over the course of development, including construction-specific frequencies and extraction-site asymmetries.
  The resulting fine-grained labels enable future work in both acquisition and computational studies, which we demonstrate with a case study using filtered corpus training with language models.
\end{abstract}

\vspace{-1pt}
\section{Introduction}
\vspace{-1pt}

Language speakers possess abstract linguistic knowledge: instead of memorizing the low-level distribution of language inputs we receive, we instead formulate knowledge of abstract structures that generalizes beyond specific string combinations.
How do children learn such abstract linguistic knowledge from limited input, and how do they generalize across similar constructions?
The \textit{argument from the poverty of the stimulus} claims that human learners are endowed with innate inductive biases which enable them to arrive at linguistic knowledge that goes beyond what is directly evidenced by the input data~\citep{chomsky1986knowledge}.
From an alternative perspective, statistical learning theories argue that input frequency plays a central role: learners extract distributional regularities from input and use them to build increasingly abstract representations over time (e.g., \citealp{rowland2003determinants}).

Filler-gap dependencies (FGDs) have long been a central topic in syntactic acquisition.
FGDs serve as a natural testbed for linguistic generalization: do learners formulate an abstract dependency between fillers and gaps 
that unifies superficially different syntactic constructions \cite{chomsky1977}, or do they begin with shallow, lexically anchored patterns and only later generalize across items and constructions \cite{Diessel2005}?

\begin{table*}[h]
  \centering
  \small
  \setlength{\tabcolsep}{4pt}
  \renewcommand{\arraystretch}{1.08}
  \resizebox{\textwidth}{!}{%
  \begin{tabular}{llll}
    \toprule
    Construction & Extraction Site & Label & Sample Sentence \\
    \midrule
    \multirow{6}{*}{Matrix wh-questions}
    & Subject & SMQ & \textbf{Who} \underline{\hspace{1em}} praised the student? \\
    & Object & OMQ & \textbf{Who} did the professor praise \underline{\hspace{1em}}? \\
    & Adjunct & AMQ & \textbf{When} did the professor praise the student \underline{\hspace{1em}}? \\
    & Polar & PMQ & Did the professor praise the student? \\
    & Plain/Fragment & PlainMQ & \textbf{Who}? \\
    & Cross-clausal & CC\_ & \textbf{Who} did the student say [that the professor praised \underline{\hspace{1em}}]? \\
    \midrule
    \multirow{4}{*}{Embedded wh-questions}
    & Subject & SEQ & I wonder [\textbf{who} \underline{\hspace{1em}} praised the student]. \\
    & Object & OEQ & I wonder [\textbf{who} the professor praised \underline{\hspace{1em}}]. \\
    & Adjunct & AEQ & I wonder [\textbf{why} the professor praised the student \underline{\hspace{1em}}]. \\
    & Polar & PEQ & I wonder [\textbf{whether} the professor praised the student]. \\
    \midrule
    \multirow{6}{*}{Relative clauses}
    & Subject & SRC & The professor [\textbf{who} \underline{\hspace{1em}} praised the student] smiled. \\
    & Object & ORC & The professor [\textbf{who} the student praised \underline{\hspace{1em}}] smiled. \\
    & Adjunct & ARC & The day [\textbf{when} the professor praised the student \underline{\hspace{1em}}] was memorable. \\
    & Possessive & PRC & The professor [\textbf{whose} student won the prize] smiled. \\
    & Subject (reduced) & SRC\_reduced & The professor [(\textbf{that/who} \underline{\hspace{1em}} was) praised by the student] smiled. \\
    & Object (reduced) & ORC\_reduced & The professor [\textbf{(that/who)} the student praised \underline{\hspace{1em}}] smiled. \\
    \bottomrule
  \end{tabular}
  }
  \caption{Target filler-gap constructions and subtypes by extraction site, with gap positions underlined.}
  \label{tab:target_constructions}
\end{table*}

Adjudicating between  the poverty of stimulus argument and statistical learning theories requires a  characterization of the data that learners are exposed to at each stage of learning (e.g.,~\citealp{pullum2002empirical, legate2002empirical}).
A major challenge to this enterprise stems from the granularity and scale that are required. FGDs can occur in a variety of sentence constructions (e.g., matrix and embedded wh-questions, relative clauses, clefting and pseudo-clefting, and topicalization), and the filler's extraction site (e.g., subject versus object positions) also plays an important role in both acquiring and processing FGDs  \cite{ambridge2015ubiquity}. 
\citet{pearl2013computational} manually annotated two corpora from the CHILDES database of child-directed speech~\citep{MacWhinney2000} with trace information that indicates filler-gap movement.
\citet{hsiao2023nature} similarly used manually annotated data to study the distributions of types of relative clauses. 
However, manual annotation is infeasible as the scale of the data increases. This issue could however  be overcome with an automated system that was sufficiently accurate (see \citet{carslaw2025automatic} for a project in this direction but for clausal embedding rather than FGDs).

In this study, we develop an automated tool for annotating FGD constructions that allows us to characterize the distribution of various FGD constructions.  Our specific contributions are:
\begin{enumerate}[topsep=2pt, itemsep=1pt, parsep=0pt]
  \item We present an automated detection tool, leveraging the strength of both constituency and dependency parsing, for identifying three well-studied FGD constructions, matrix wh-questions, embedded wh-questions, and relative clauses, each subtyped by extraction site.



  \item We apply our tool to 57 English corpora from the CHILDES database and present descriptive statistical analyses of the FGD distribution in both child-directed speech and children's production. The resulting dataset and the detector program is released on \href{https://github.com/herbert-zhou/filler_gap_detector_childes.git}{\faGithub \ Github}.\footnote{The code base is available at: \url{https://github.com/herbert-zhou/filler_gap_detector_childes.git}}

  \item We also outline how large datasets of FGDs with fine-grained annotations can be fruitfully applied both to address open questions in the acquisition literature as well as to study linguistic generalization in modern language models using methods such as filtered corpus training and input attribution.
\end{enumerate}
\vspace{-1pt}
\section{Target Constructions} \label{sec:target_constructions}
\vspace{-1pt}

We focus on three core \textbf{constructions}---matrix wh-questions, embedded wh-questions, and relative clauses---all of which instantiate the hallmark of filler-gap dependencies: a leftwardly displaced filler (typically a wh-phrase or relative operator) that is interpreted in a position (the gap) where no pronounced constituent appears.
Within each construction, we distinguish subtypes by \textbf{extraction site} (i.e., subject vs.\ object vs.\ adjunct), as these distinctions often correlate with distributional and processing differences and therefore are relevant for characterizing learners' input. This adds another level of granularity that no previous automated approaches have achieved.
Table~\ref{tab:target_constructions} summarizes the construction types and extraction-site subtypes we target, with fillers highlighted and gap positions explicitly marked by an underline to make the dependency visible in the surface string.\footnote{We excluded certain constructions that are closely related to some shown in Table~\ref{tab:target_constructions},
such as (i) \textbf{free relatives} (e.g., \textit{I read what the professor wrote})
and (ii) \textbf{infinitival relative clauses} (e.g., \textit{The professor picked a time for students to come}).
}
Although polar matrix questions do not involve wh-movement, they belong to the matrix question family and could affect generalization --- it is potentially of interest to analyze whether PMQs behave like matrix wh-questions, and whether and how children generalize across PMQs and other question types. We thus included PMQs as one of our target constructions.

\vspace{-1pt}
\section{Detection Algorithms}\label{sec:detection_algo}
\vspace{-1pt}

We take a hybrid approach to identifying the presence and type of FGDs that combines the strengths of constituency and dependency parsing. We use the spaCy dependency parser~\citep{Honnibal_spaCy_Industrial-strength_Natural_2020} and the spaCy implementation of the Berkeley Neural parser (\citealp{kitaev-klein-2018-constituency}; \citealp{kitaev-etal-2019-multilingual}), a widely used constituency parser based on a self-attentive architecture. Here we present the core steps of the detection algorithms and demonstrate the necessity of using both parse types.

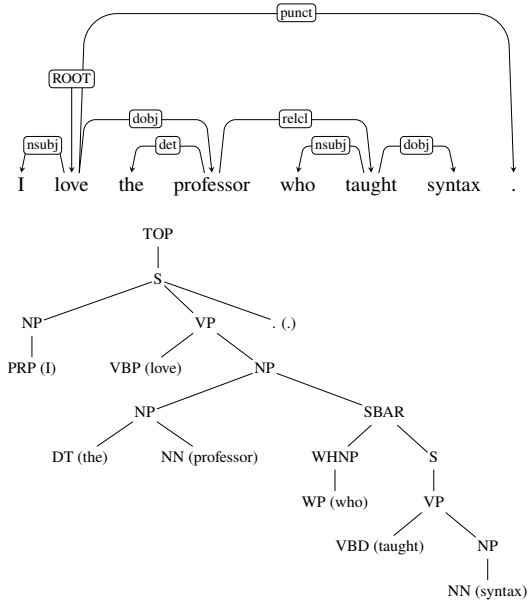
\begin{figure}[t]
  \centering
  \resizebox{0.9\columnwidth}{!}{\input{trees/dep_rc_demo}}\par\medskip
  \resizebox{0.9\columnwidth}{!}{\input{trees/cons_rc_demo}}
  \caption{Dependency (top) and constituency (bottom) parses of a sentence containing a subject relative clause.}
  \label{fig:rc_demo}
\end{figure}

\subsection{Detection Steps}
Our detection algorithm consists of three detectors corresponding to the three constructions, each implementing a set of heuristics for construction detection and extraction site subtyping. 
Since a complex sentence can involve multiple constructions and subtypes, our algorithm applies all three detectors separately and outputs a list of all detected categories from Table~\ref{tab:target_constructions}.  
In this section, we illustrate core detection steps for relative clauses (see Appendix~\ref{app:rc_detection} for detailed elaboration), which can be generalized to other constructions. For the complete algorithm, we refer readers to the released code base, as it is infeasible to present it fully here.

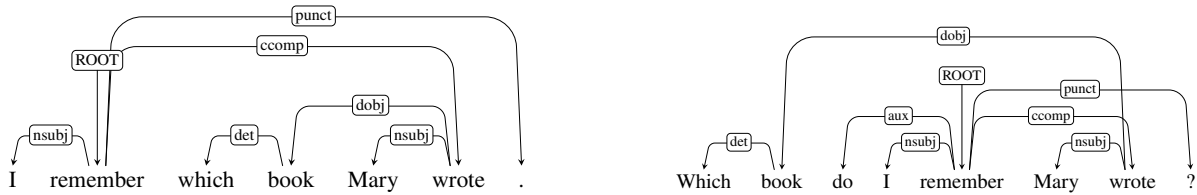
\begin{figure*}[t]
  \centering
  \begin{minipage}[t]{0.45\textwidth}
    \centering
    \resizebox{\linewidth}{!}{\input{trees/dep_embed.tex}}
  \end{minipage}\hfill
  \begin{minipage}[t]{0.45\textwidth}
    \centering
    \resizebox{\linewidth}{!}{\input{trees/dep_matrix.tex}}
  \end{minipage}
  \caption{Dependency parses of an embedded question (left) and a matrix question (right).}
  \label{fig:dep_questions}
\end{figure*}

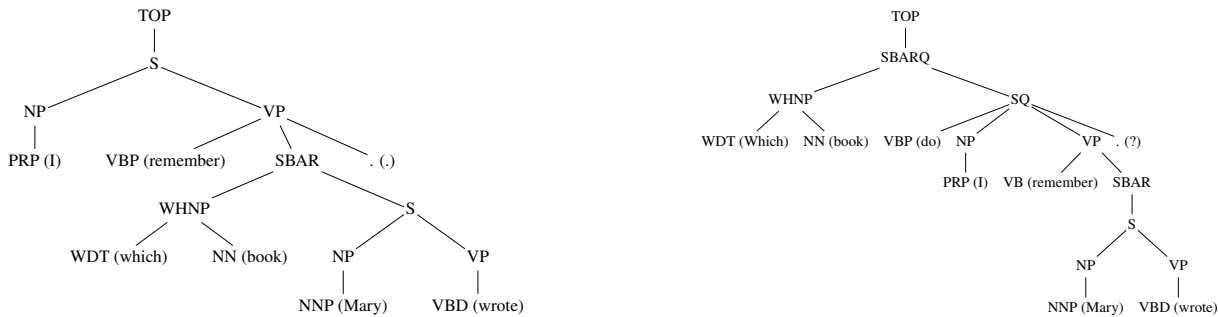
\begin{figure*}[t]
  \centering
  \begin{minipage}[t]{0.43\textwidth}
    \centering
    \resizebox{\linewidth}{!}{\input{trees/cons_embed.tex}}
  \end{minipage}\hfill
  \begin{minipage}[t]{0.43\textwidth}
    \centering
    \resizebox{\linewidth}{!}{\input{trees/cons_matrix.tex}}
  \end{minipage}
  \caption{Constituency parses of an embedded question (left) and a matrix question (right).}
  \label{fig:cons_questions}
\end{figure*}

\vspace{-1pt}
\paragraph{Step 1: Core Structure Detection} Starting with the constituency parse of the input sentence, illustrated in Figure~\ref{fig:rc_demo}, we recursively detect all occurrences of the local structural signature for relative clauses, \texttt{NP -> NP SBAR}. The \texttt{NP} child node denotes the noun phrase modified by the relative clause \texttt{SBAR}. 

\vspace{-1pt}
\paragraph{Step 2: Wh-category Identification} For each detected \texttt{NP -> NP SBAR}, we look for a wh-category immediately dominated by \texttt{SBAR} and retrieve the entire span of the wh-phrase, in this case \texttt{WHNP} (\textit{who}). For sentences where the wh-phrase is omitted, such as reduced relative clauses, we branch into a separate set of heuristics for detecting their properties using constituency parses: the absence of a wh-word makes dependency relations less robust.

\vspace{-1pt}
\paragraph{Step 3: Extraction Site Inference} To identify the extraction site, we look for the lowest \texttt{S} node under \texttt{SBAR} (to tolerate intervening material) and check whether there is an \texttt{NP} preceding a \texttt{VP}. If that is the case and the wh-word is \texttt{WHNP}, then it is likely that the extraction is from object position, as the subject position is occupied by another element.  If instead no \texttt{NP} precedes \texttt{VP} below \texttt{S} (as in Figure~\ref{fig:cons_questions}), this suggests that extraction has taken place from subject position. Adjunct categories like \texttt{WHPP} and \texttt{WHADVP} follow similar logic of checking the existence of an \texttt{NP} at subject positions.

\vspace{-1pt}
\paragraph{Step 4: Dependency Validation} We verify the labels hypothesized from the constituency parse with the dependency parse. We  start by verifying that the wh-word that was identified below the  detected \texttt{SBAR}  is dependent on the verb inside the relative clause (\textit{taught}) with the relation \texttt{relcl}.
In cases like Figure~\ref{fig:cons_questions} where constituency heuristics labeled the sentence as containing a subject relative clause, we confirm that  the wh-word has either an \texttt{nsubj} or \texttt{nsubjpass} dependency relation. Finally, we verify the relation between \textit{taught} and the modified noun by checking if the noun that \textit{taught} depends on is the noun in the \texttt{NP} from the constituency parse.

Additional construction-specific heuristics are added to deal with noise, parsing errors, and constructions that we intentionally omitted (see Appendix~\ref{app:embQ} for details). We note that there is unlikely to be a perfect solution that could deal with all edge cases, since adding more heuristics that fix one set of cases will lead to underdetection or overdetection in another set of cases.

\subsection{Arguments for a hybrid approach}\label{subsec:hybriding}

Constituency parsing and dependency parsing offer two ways of characterizing  syntactic structure: constituency structures make clausal boundaries and complement types explicit, which is useful for detecting the construction types; dependency relations provide more direct access to head-dependent configurations that are useful for identifying extraction sites.
While it is possible to analyze the majority of sentences using only one parsing strategy, there are situations where information offered by one parser is systematically insufficient and could lead to false detection. In contrast, combining information from both leads to more robust detection.

Consider the subject matrix question \textit{Which book do I remember Mary wrote?} and its embedded question counterpart \textit{I remember which book Mary wrote.}
As shown in Figure~\ref{fig:dep_questions}, both sentences exhibit superficially identical dependency relations (except for the extra auxiliary \textit{do} in the matrix question).
In particular, the embedded wh-clause in the matrix question is not distinguished from a genuine embedded question solely from the dependency structure --- the dependency parse does not encode the scope of the wh-operator.\footnote{One could attempt to distinguish between these using the linear position of the wh-word, but doing so will introduce other complexities and ignores the natural structural generalization.}
On the other hand, in Figure~\ref{fig:cons_questions}, the \texttt{VP -> VP SBAR} structure clearly marks embedded questions and the \texttt{SBARQ -> WHNP SQ} structure marks matrix questions.
This is an example of utilizing additional information provided by constituency parsing to resolve the ambiguity between matrix and embedded questions that dependency parsing alone cannot adjudicate.
See Appendix~\ref{app:dep_cons_complementary} for an example of how dependency parsing complements constituency parsing. An additional practical motivation is imperfections in the source data: child speech can be noisy, ungrammatical, and subject to transcription errors, which could lead to parsing errors. Combining information from both parses makes the detection more robust to noise.

\vspace{-6pt}
\section{Evaluation}
\vspace{-6pt}

To assess the effectiveness of our detection algorithm, we first conducted a small-scale, manual evaluation of the labels produced by our approach. We then compared our detection to information from a human-annotated corpus \citep{pearl2013computational}.

\subsection{Manual Evaluation}
\vspace{-4pt}
As a sanity check, we first applied our detector to CHILDES and sampled 100 sentences from both child-directed speech and child speech that the detector identified as belonging to one of six core categories: the three FGD constructions, each with extraction sites of subject or object. 
Five human annotators with expertise in linguistics (faculty and students in linguistics) then provided binary assessments of whether each sentence contains the labeled construction. 
The detector was taken to have been accurate for a given sentence if a majority of the human judges responded positively to the assigned label. Results are shown in Table~\ref{tab:manual_precision}. 
Our detector achieved nearly perfect accuracy for child-directed speech and somewhat lower---but still fairly high---accuracy for child speech. Manual inspection of error cases indicated that child speech includes more edge cases due to ungrammatical, fragmented speech as well as transcription errors, which led to lower precision compared to child-directed speech.


\begin{table}[h]
  \centering
  \small
  \setlength{\tabcolsep}{5pt}
  \begin{tabular}{lcc}
    \toprule
    Category & {Child-directed} & {Child speech} \\
    \midrule
    SMQ & 1.00 & 0.83 \\
    OMQ & 0.99 & 0.89 \\
    \midrule
    SEQ & 0.95 & 0.95 \\
    OEQ & 0.94 & 0.97 \\
    \midrule
    SRC & 0.98 & 0.98 \\
    SRC reduced & 0.98 & 0.92 \\
    ORC & 1.00 & 0.91 \\
    ORC reduced & 0.95 & 0.80 \\
    \bottomrule
  \end{tabular}
  \caption{Human annotated detector accuracy by construction and gap site (out of 100 sentences per category) for child-directed speech and child speech.}
  \label{tab:manual_precision}
\end{table}
\vspace{-3pt}

\vspace{-5pt}
\subsection{Comparison with \citeauthor{pearl2013computational}}
\vspace{-3pt}

We next performed a larger scale evaluation by comparing our detection results with human annotations of FGDs from \citet{pearl2013computational}. 
This corpus includes annotations of 56,461 child-directed utterances from the Brown corpus (\citealp{brown1973first}) and the Valian corpus~\citep{valian1991syntactic} with trace and coindexing information represented in constituency parses. This information allowed us to infer the construction and subtype information present in the human annotations.\footnote{See Appendix~\ref{app:ps13_details} for details on their annotation schema and our inference method.} 

We run our detector on the same corpus and use the human annotations as gold labels. This allows us to compute precision (how often our detector is correct when claiming a label) and recall (how often our detector claims a label given that human annotation produced that label). These results are shown in Table~\ref{tab:ps13_stats}, together with the total number of sentences of each category.\footnote{Since our detector provides more categories (see Table~\ref{tab:target_constructions} for the set of labels we consider) than what we inferred from tree annotations, we merged secondary categories with their main categories: e.g., those labeled as cross-clausal SMQs were treated as SMQs when computing the precision and recall.}
We omitted PMQs in this comparision since they do not involve wh-movement and lack trace labels in human annotations, and we omitted PRCs since none occurred in human annotations. We also omitted reduced SRC and ORC because we found inconsistency in trace annotations.\footnote{We note that for most of the categories for embedded questions and relative clauses, the total numbers of detected sentences for recall are greater than those for precision. This is because human annotations include sentences that our detector  omitted by design (elaborated on in Section~\ref{sec:target_constructions}). For sentences labeled by human annotations but not detected by our detector, we manually checked and counted them as true positives if they were indeed among those omitted constructions.}
All labels except for adjunct relative clauses achieved F1 scores above 0.8, with more than half above 0.9. No category had extremely unbalanced precision and recall, suggesting that our detector largely aligns with human annotations for deciding constructions and extraction sites. Given both evaluation results, we conclude that despite being imperfect, our automated detector is a viable solution for large-scale, fine-grained detection of our target structures.

\begin{table}[t]
  \centering
  \small
  \setlength{\tabcolsep}{4pt}
  \renewcommand{\arraystretch}{1.05}
  \resizebox{\columnwidth}{!}{%
    \begin{tabular}{lccc}
      \toprule
      Category & Precision (total) & Recall (total) & F1 \\
      \midrule
      SMQ & 0.827 (712) & 0.792 (716) & \textbf{0.809} \\
      OMQ & 0.908 (4950) & 0.977 (4464) & \textbf{0.941} \\
      AMQ & 0.921 (1839) & 0.957 (1746) & \textbf{0.939} \\
      \midrule
      SEQ & 0.925 (146) & 0.894 (236) & \textbf{0.909} \\
      OEQ & 0.958 (642) & 0.895 (1325) & \textbf{0.925} \\
      AEQ & 0.808 (339) & 0.905 (924) & \textbf{0.854} \\
      PEQ & 1.000 (243) & 0.905 (611) & \textbf{0.950} \\
      \midrule
      SRC & 0.924 (157) & 0.883 (247) & \textbf{0.903} \\
      ORC & 0.901 (121) & 0.810 (473) & \textbf{0.853} \\
      ARC & 0.842 (38) & 0.713 (94) & 0.772 \\
      \bottomrule
    \end{tabular}
  }
  \caption{Precision (against parser labels), recall (against annotation labels), and F1 scores (bolded if $> 0.8$) by construction and extraction site.}
  \label{tab:ps13_stats}
\end{table}

\begin{figure*}[t]
  \centering
  \includegraphics[width=\textwidth]{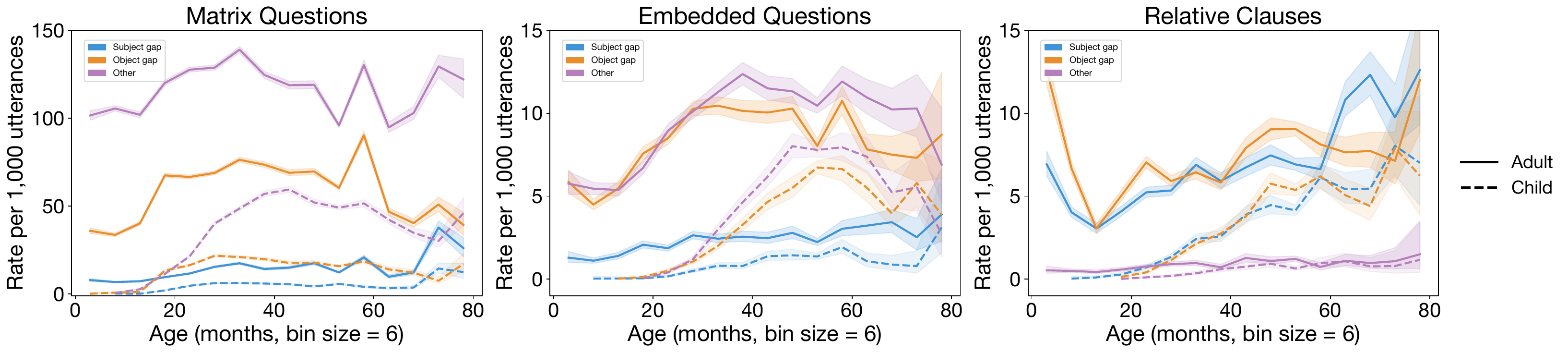}
  \caption{Adult (solid) and child (dashed) speech distributions across time by constructions and extraction sites. Uncertainty is shown as 95\% Wilson intervals (wider in sparser bins).}
  \label{fig:overall_triplots}
\end{figure*}
\vspace{-3pt}

\begin{figure*}[t]
  \centering
  \includegraphics[width=0.85\textwidth]{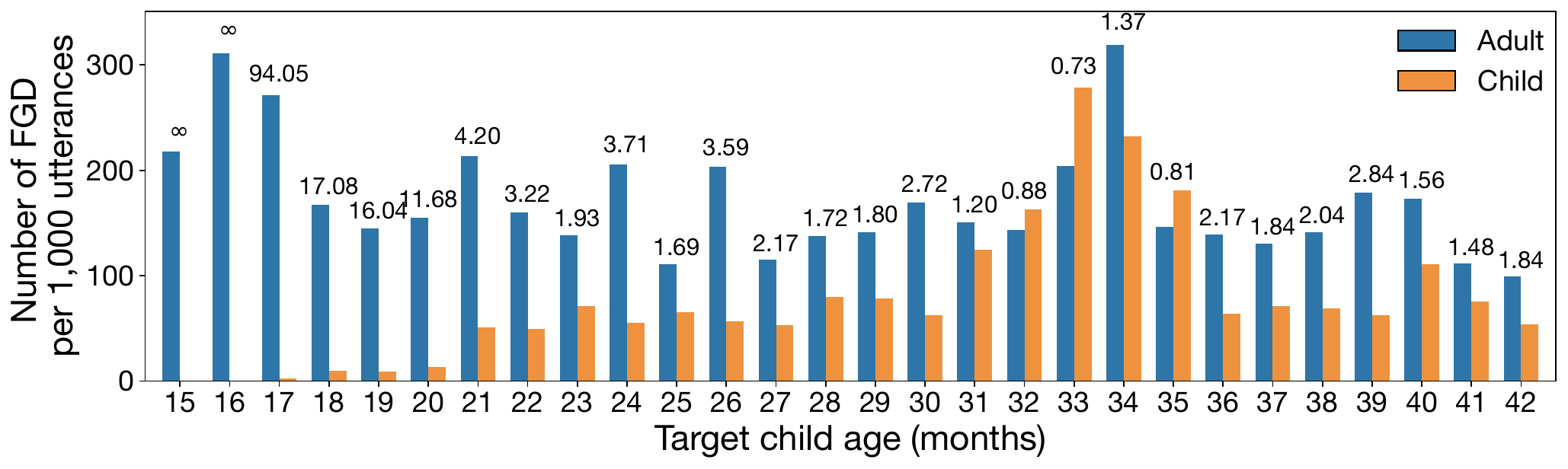}
  \caption{Per-1,000 utterance rates of all filler-gap dependency sentences received and produced by Laura.}
  \label{fig:laura_by_month}
\end{figure*}

\vspace{-1pt}
\section{A Case Study on CHILDES}
\vspace{-1pt}

CHILDES \cite{MacWhinney2000} is the largest publicly available collection of longitudinal, naturalistic child and child-directed speech, making it uniquely suitable for quantifying the input distributions that could shed light on children’s acquisition of filler-gap dependencies.
By applying our detection algorithm to CHILDES, we move beyond small-scale, hand-annotated samples to get a better measure of how often different dependency types occur in both child and adult  speech,\footnote{In the rest of this paper, we use the term \textit{adult speech} instead of child-directed speech, although we acknowledge that not all child-directed speech is produced by adults.} how these frequencies change with child age, and how they differ across extraction sites, which are known to modulate both learning and processing (see Section~\ref{sec:future} for further discussion).
These corpus-scale measurements serve two complementary goals: they provide a descriptive picture of what children hear and produce across time, and they yield construction-specific summaries that can be used to evaluate acquisition theories and to design computational studies with explicit input control, further discussed in Section~\ref{sec:computational}.
These goals are hard to achieve with the existing human-annotated corpora from \citet{pearl2013computational} since they only includes child-directed speech and cover a sparse age range.

We accessed the entire English-NA CHILDES database via \texttt{childes-db}~\citep{sanchez2019childes}, extracting 3,194,544 utterances distributed over 50,327 chat session transcripts from 57 corpora.
Although target children's ages ranged up to 144 months, we decided to focus on utterances with target children in the range 3 to 80 months old since data outside this range was relatively sparse.
After removing utterances missing age information, we retained 2,841,084 utterances (92.42\% of total utterances) for further analyses.
See Appendix~\ref{app:childes_processing} for more details of our data pre-processing.

\vspace{-1pt}
\subsection{Distributions Across Time}

Understanding adult input and child output distributions over the course of development is central for evaluating acquisition theories, which differ in whether children’s generalizations are driven by input frequency and in whether child production tracks input in construction-specific ways. We grouped utterances into 6-month age bins and, for each construction subtype, computed rates per 1,000 utterances for adult and child speech.

Figure~\ref{fig:overall_triplots} shows three robust patterns. 
First, matrix questions are roughly an order of magnitude more frequent than embedded questions and relative clauses (note the different y-axis scales), consistent with their lower structural complexity. 
Second, for both wh-question families, adjunct and polar questions outnumber subject and object extractions, and object extractions consistently exceed subject extractions. For relative clauses, subject and object extractions are comparable and both outnumber adjunct relatives. 
Third, for both wh-question families across ages, child production closely mirrors adult production in the rankings of the extraction sites.
For relative clauses, subject and object RCs show a steady increase through about 55 months.


\begin{figure*}[t]
  \centering
  \includegraphics[width=0.85\textwidth]{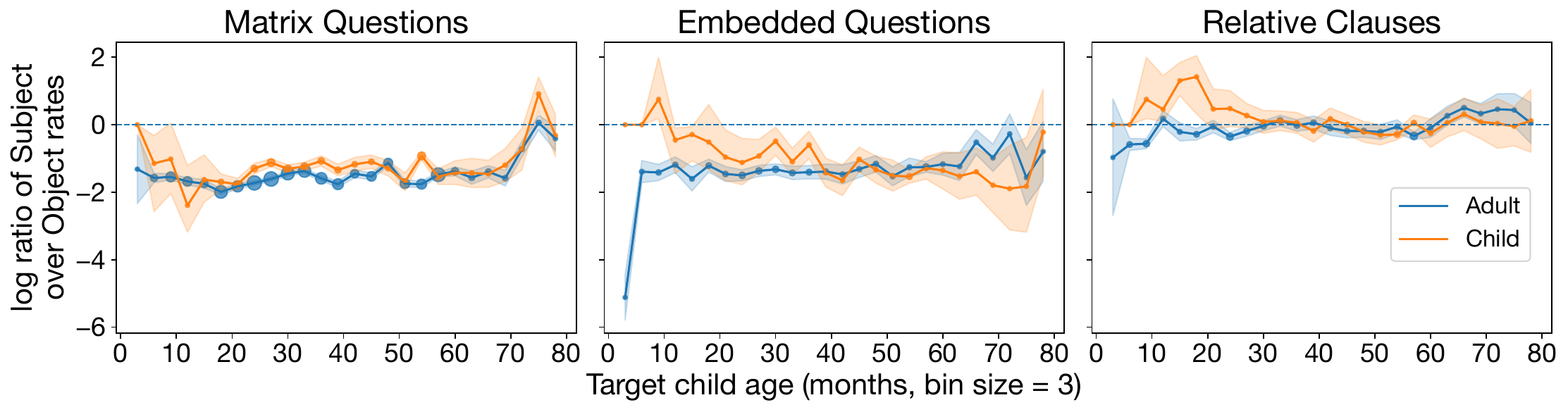}
  \caption{Subject- versus object-extracted log ratios for each construction across ages, binned by 3 months. Sizes of dot shades indicate number of total utterances within each bin.}
  \label{fig:sor_plot}
\end{figure*}

\vspace{-4pt}
\subsection{Total Exposures of One Individual Child}
\vspace{-2pt}

How much relevant input does an individual child receive over development, and what share do filler-gap constructions occupy? 
These questions matter for acquisition theories and for evaluating how human-like LLM training inputs are (see Section~\ref{sec:computational}). As a reference point, we report one longitudinal case study as a reference for future work.

\begin{figure}[t]
  \centering
  \includegraphics[width=\columnwidth]{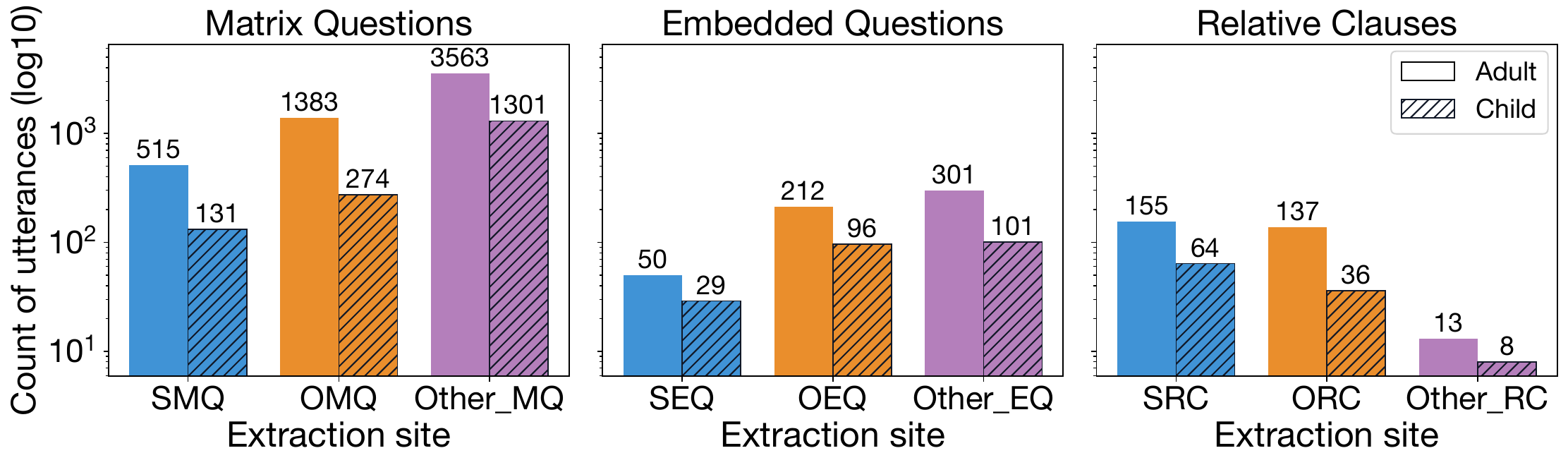}
  \caption{Number of utterances with targeted constructions received and produced by Laura in log scale.}
  \label{fig:laura_total}
\end{figure}

Among longitudinal English CHILDES corpora, Laura in the Braunwald corpus~\citep{braunwald1971mother} has the largest amount of data: 75,740 adult and child utterances across 900 transcripts, spanning the age range of 15-77 months. We focus on the period of 15 to 42 months because later months are sparse. 
Figure~\ref{fig:laura_total} plots absolute counts of each FGD constructions on a log scale, with totals labeled.
The resulting profile mirrors the global trends in Figure~\ref{fig:overall_triplots}: matrix questions dominate embedded questions and relative clauses, and object extractions are consistently more frequent than subject extractions for wh-questions but not for relative clauses.

To contextualize these counts within overall exposure, Figure~\ref{fig:laura_by_month} shows the by-month proportion of utterances containing any FGD among all utterances, normalized as per-1,000-utterance rates. Each month is annotated with the adult versus child proportion ratio (values $>1$ indicate higher FGD prevalence in adult speech). Despite substantial month-to-month variation in total utterances, FGD utterances remain a relatively small fraction in both streams. Laura’s first detected FGD occurs at 17 months, and the adult versus child ratio decreases with age, dropping below 1 between 32-35 months, indicating that Laura’s speech contains a higher proportion of FGD utterances than adult speech during that period. 
This increase might be taken to indicate a gradual acquisition of FGD constructions, but it is also consistent with a gradual increase in the use of sentences with higher syntactic complexity.

While longitudinal corpora cannot capture all utterances a child receives or produces, this case study provides stable \textit{relative} distributions over constructions and extraction sites. These summaries can be interpreted as a scaled estimate of children’s exposure and serve as an empirical target for acquisition theories and for computational modeling with explicit input control \cite{frank2023bridging}.

\vspace{-2pt}
\subsection{Subject versus Object Gaps}\label{subsec:so_ratio}
\vspace{-2pt}

\begin{figure}[t]
  \centering
  \includegraphics[width=0.95\columnwidth]{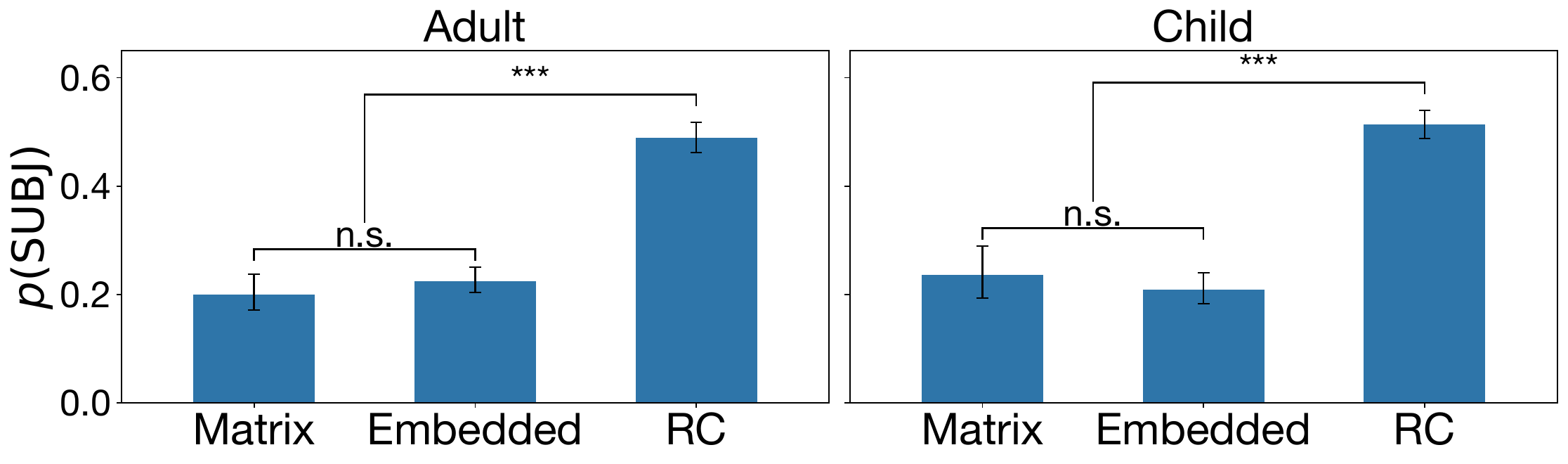}
  \caption{Cross-construction comparisons of subject share across age.}
  \label{fig:p_subj_significance}
\end{figure}
\vspace{-2pt}

A central question in FGD acquisition is generalization: to what extent can children generalize knowledge of a construction to other related constructions? Extraction site is a natural case to test generalization. For instance, \citet{gagliardi2016discontinuous} studied whether gap sites show systematic extraction biases, and whether such biases differ by construction (see Section~\ref{sec:future} for more discussion). 
We focus on subject- vs.\ object-extracted utterances as a first step.
For each construction and age bin, we computed the log ratio of the rate-per-1,000 utterances for subject versus object extractions,
$\log\frac{\#\textsc{subj}+\epsilon}{\#\textsc{obj}+\epsilon}$, where $\#\textsc{subj}$ denotes the number of subject-extracted utterances per 1,000 utterances. 
Values above 0 indicate a subject bias and values below 0 an object bias. 
Figure~\ref{fig:sor_plot} shows a construction-specific profile: matrix questions are strongly object-biased in both adult input and child production; embedded questions exhibit a similar object bias; and relative clauses are closer to being balanced. Consistent with Figure~\ref{fig:overall_triplots}, child trajectories broadly track adult trajectories within each family, suggesting that child speech preserves both the direction and relative strength of the input bias.

To compare biases across constructions, we computed the subject share
$p(\textsc{subj})=\frac{\#\textsc{subj}}{\#\textsc{subj}+\#\textsc{obj}}$ for each age bin and each construction. We then averaged across bins that meet a minimum-count threshold, with pairwise comparison across constructions shown in Figure~\ref{fig:p_subj_significance}. We found no significant difference in $p_{\textsc{subj}}$ between matrix and embedded questions, with both being below 0.5 for adult and child speech; whereas relative clauses have a reliably higher subject share than both question types, with $p_{\textsc{subj}} \approx 0.5$.\footnote{See Table~\ref{tab:delta_subj_summary} in Appendix~\ref{app:more_individual_results} for details.}
Together, these results suggest that extraction biases are not uniform across filler-gap constructions: these biases are similar across question types, but relative clauses differ from questions.


\vspace{-2pt}
\section{Linguistic Generalization in BabyLMs} \label{sec:computational}
\vspace{-2pt}
\subsection{Motivation}
Beyond analyses of human acquisition, our detector also supports targeted tests of linguistic generalization in modern language models (LMs): to what extent do human-scale LMs generalize across related constructions, and how does the input affect such generalization? 
Since our detector partitions data by construction type and extraction site, it enables controlled interventions on training corpora that can help to answer such questions.

As a case study in this direction, we used the filtered corpus training paradigm, in which a model is trained on data with selected constructions removed to test whether generalization to held-out patterns depends on direct exposure or transfers across related constructions \cite{jumelet2021language,misra2024language,patil2024filtered}. 
Prior work has applied analogous ideas to filler-gap phenomena in restricted settings:
\citet{howitt2024generalizations} manipulate exposure via targeted augmentation (e.g., clefting and topicalization) and interpret improvements elsewhere as evidence for shared representations; 
\citet{lan2024large} extend augmentation to rarer movement phenomena (e.g., parasitic gaps and across-the-board movement) and find improved generalization;
\citet{chang2025mind} evaluated existing models on a suite of tests testing a broad range of constructions and reported that enriching the dataset with filler-gap dependencies leads to improvements in evaluations but does not lead to human-level performance. 
There has also been work characterizing the abstract causal mechanisms in LMs responsible for processing filler-gap dependencies \cite{boguraev2025causal, desai2026fillingmechanismslmslearn}. Our detector makes training-data interventions scalable across larger corpora, more construction families, and extraction-site subtypes, enabling more precise tests of generalization.

\subsection{Experiment Setup}
\paragraph{Corpus Filtering}
We conducted filtered corpus training on relatively small-scale LMs trained on child-directed language,  
specifically the CHILDES component from the 10-million-token version of the BabyLM dataset\footnote{The dataset was accessed from \url{https://babylm.github.io/}} \citep{warstadt2023findings}. We used only child-directed speech as training input in order to simulate the types of sentences encountered during acquisition. This results in a total of 360,146 sentences (2,091,023 tokens). We experiment with filtering each of the three constructions, resulting in three conditions, one for each construction. 
The number of sentences and tokens \textit{removed} for each filtered condition is given in Table~\ref{tab:filtered_data}.
\begin{table}[t]
  \centering
  \small
  \setlength{\tabcolsep}{5pt}
  \begin{tabular}{ccc}
    \toprule
    Filtered Construction & \# Sentences & \# Tokens \\
    \midrule
    Matrix questions & 69,214 & 502,555 \\
    Embedded questions & 3,765 & 51,930 \\
    Relative clauses & 5,850 & 62,637 \\
    \bottomrule
  \end{tabular}
  \caption{Numbers of tokens and sentences removed from the training corpus for each construction family.}
  \label{tab:filtered_data}
\end{table}
\noindent
For each construction, we also produced a control dataset 
by removing the same number of sentences but randomly selected instead of targeting a specific construction.
To ensure that the ablated and control datasets had the same number of sentences and comparable length distributions, we removed the same number of sentences in both cases and matched the removed sentences in number of tokens. 
This control condition allows us to attribute differences between models trained on the ablated versus control datasets specifically to the absence of the ablated construction, rather than to reduced training size or sentence length.


\paragraph{Evaluation Dataset}
To assess models' FGD knowledge, we evaluated models with 3432 synthetically constructed minimal pairs for matrix questions and 5000 synthetically constructed minimal pairs for each of embedded questions and relative clauses.
Sentences in each pair differ in whether they contain a gap, and they share the same continuation after the (filled or unfilled) gap position, as shown in Table~\ref{tab:evaluation}. 
We compared the probability of the continuation conditioned on the contexts of the two sentences and made binary judgments on whether the grammatical sentence has a higher continuation probability than the ungrammatical one.
We constructed $15$ templates and created the minimal pairs by filling in the templates with high frequency lexical items selected from the BabyLM dataset.\footnote{We did not evaluate on subject relative clauses because the format of minimal pairs used with object relatives would not lead to a grammaticality difference in the case of subject relative clauses.} We manually checked their selectional restrictions to ensure semantic naturalness. 
See Appendix~\ref{app:babylm_eval} for details of the evaluation dataset.

\begin{table}[h]
  \centering
  \small
  \setlength{\tabcolsep}{4pt}
  \renewcommand{\arraystretch}{1.12}
  \resizebox{0.78\columnwidth}{!}{%
  \begin{tabularx}{\columnwidth}{p{0.35\columnwidth}X}
    \toprule
    \textbf{Construction $\times$ Extraction Site} & \textbf{Sample Evaluation Pairs} \\
    \midrule

    \begin{tabular}[t]{@{}l@{}}
      Matrix Question\\
      Object Gap
    \end{tabular}
    &
    \begin{tabular}[t]{@{}l@{}}
      \textbf{What} will you build \underline{\hspace{1em}} \textit{today} \\
      *You will build \underline{\hspace{1em}} \textit{today} \\
      \\[-0.4em]
      You will build \textit{it} \\
      *\textbf{What} will you build \textit{it}
    \end{tabular}
    \\

    \midrule

    \begin{tabular}[t]{@{}l@{}}
      Matrix Question\\
      Subject Gap
    \end{tabular}
    &
    \begin{tabular}[t]{@{}l@{}}
      \textbf{Who} will \underline{\hspace{1em}} \textit{chase the doctor} \\
      *\underline{\hspace{1em}} will \textit{chase the doctor} \\
      \\[-0.4em]
      It will \textit{chase the doctor} \\
      *\textbf{Who} will it \textit{chase the doctor}
    \end{tabular}
    \\

    \midrule

    \begin{tabular}[t]{@{}l@{}}
      Embedded Question\\
      Object Gap
    \end{tabular}
    &
    \begin{tabular}[t]{@{}l@{}}
      I knew \textbf{what} you built \underline{\hspace{1em}} \textit{today} \\
      *I knew that you built \underline{\hspace{1em}} \textit{today} \\
      \\[-0.4em]
      I knew that you built \textit{it} \\
      *I knew \textbf{what} you built \textit{it}
    \end{tabular}
    \\

    \midrule

    \begin{tabular}[t]{@{}l@{}}
      Embedded Question\\
      Subject Gap
    \end{tabular}
    &
    \begin{tabular}[t]{@{}l@{}}
      I knew \textbf{who} \underline{\hspace{1em}} \textit{chased the doctor} \\
      *I knew that \underline{\hspace{1em}} \textit{chased the doctor} \\
      \\[-0.4em]
      I knew that they \textit{chased the doctor} \\
      *I knew \textbf{who} they \textit{chased the doctor}
    \end{tabular}
    \\

    \midrule

    \begin{tabular}[t]{@{}l@{}}
      Relative Clause\\
      Object Gap
    \end{tabular}
    &
    \begin{tabular}[t]{@{}l@{}}
      I knew the cake \textbf{that} \textit{you made \underline{\hspace{1em}}} \\
      *I knew that \textit{you made \underline{\hspace{1em}}} \\
      \\[-0.4em]
      I knew \textbf{that} \textit{you made it} \\
      *I knew the cake \textbf{that} \textit{you made it}
    \end{tabular}
    \\

    \bottomrule
  \end{tabularx}
  }
  \caption{Example minimal pairs. Gap positions are marked by underline and continuations marked by italics.}
  \label{tab:evaluation}
\end{table}

\subsection{Results}
\begin{figure}[t]
  \centering
  \includegraphics[width=0.86\columnwidth]{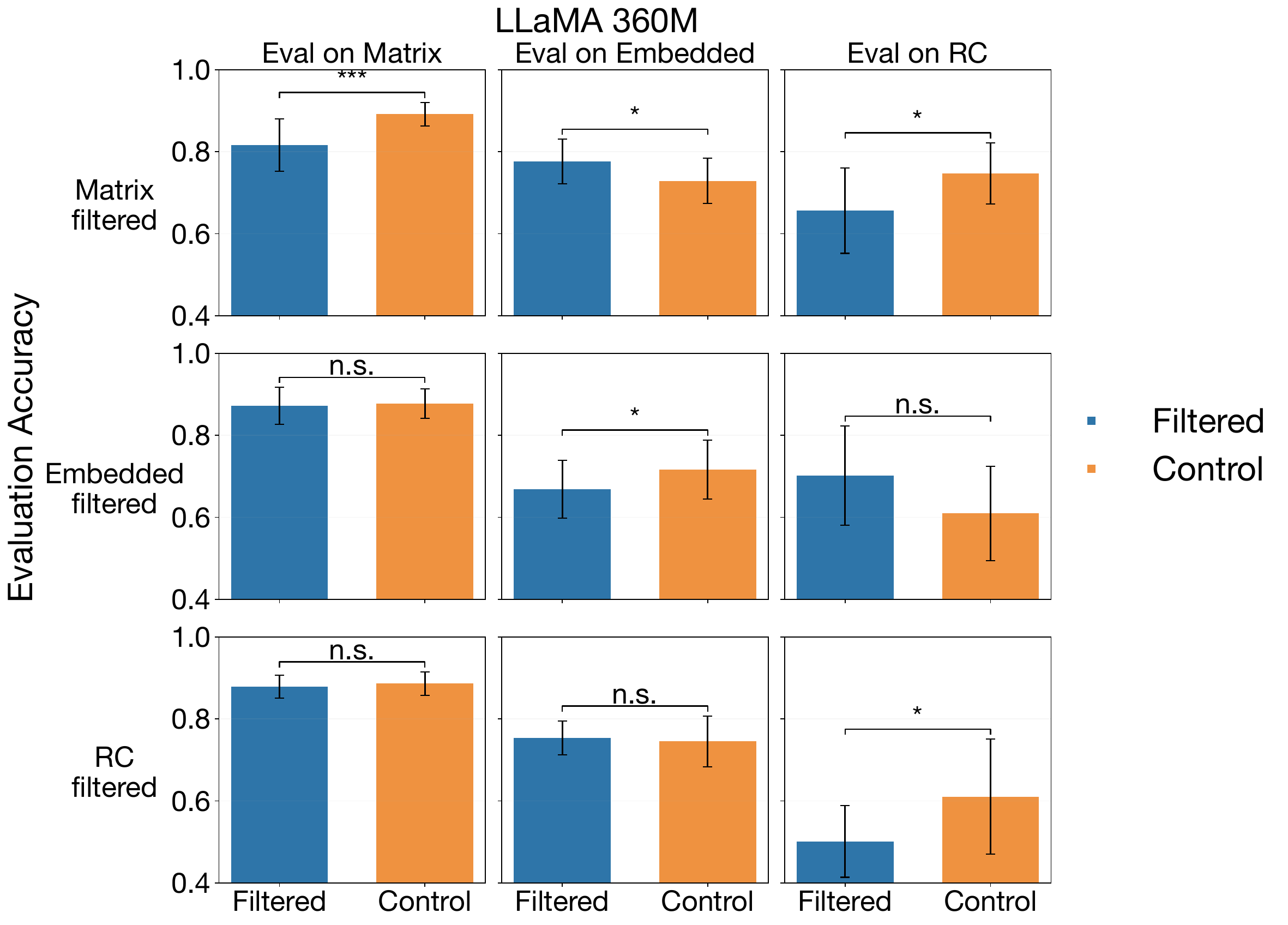}
  \caption{Cross-construction evaluations on filtered corpus training results of 15 Llama models. Evaluation accuracy means accuracy on the minimal pair tests. }
  \label{fig:babylm}
\end{figure}

For each ablated dataset and control dataset, we trained language model instances with Llama~\citep{touvron2023llamaopenefficientfoundation} and GPT-2~\citep{radford2019language} architectures with 15 different random seeds for each (see Appendix~\ref{app:babylm} for more details on training). 
Results for the Llama models are shown in Figure~\ref{fig:babylm}: compared to control models (trained on randomly filtered data), filtering each construction leads to a significant decrease in performance on the same construction, suggesting that the training input of the constructions is crucial to the acquisition of FGD knowledge. We further found that filtering matrix questions leads to significant degradation of performance for embedded questions and relative clauses, while filtering out the other two constructions did not lead to significant degradation in the performance on other constructions. Our results support the claim in \citet{boguraev2025causal} that more frequent constructions serve as sources of generalization to less frequent constructions, but not vice versa. Future studies can extend this experiment and investigate the more fine-grained generalization across extraction sites.

\vspace{-5pt}
\section{Prospective Applications}
\vspace{-5pt}
\label{sec:future}

\paragraph{Testing human acquisition theories}
Work on FGD acquisition has emphasized two questions about what children learn from input. 
One line of research contrasts complexity-based acquisition orders with frequency-based accounts: classic proposals link earlier mastery to lower linguistic complexity (e.g., wh-pronominals before wh-sententials; general verbs before more restrictive ones; \citealp{bloom1982wh}), whereas distributional approaches argue that acquisition is better predicted by how often children encounter the relevant patterns (e.g., wh-word-verb combinations; \citealp{rowland2003determinants}). 
Crucially, frequency effects are expected only when frequency is measured at the appropriate level of generalization (lexical items vs.\ constructions vs.\ subtype divisions; \citealp{ambridge2015ubiquity}). Relatedly, a second line asks which dimensions matter for characterizing FGDs and whether different dependency types share a common developmental profile: subject-object asymmetries can differ in frequency, timing, and processing difficulty (\citealp{ambridge2015ubiquity,atkinson2018developing}), and direct comparisons between wh-questions and relative clauses report dissociations across ages and roles (\citealp{gagliardi2016discontinuous,sprouse2016experimental}).

Our detector makes these debates testable at scale. It enables frequency-based analyses at the construction-by-gap-site level, precisely where prior work predicts frequency should matter \citep{ambridge2015ubiquity}. It also supports large-scale comparisons of role-sensitive trajectories across dependency types. By avoiding subtype collapse, the resulting measurements allow sharper replications and more discriminative evaluations of competing acquisition accounts. Furthermore, future work could consider extracting lexical information (e.g., wh-word, head verb, modified nouns, etc.) and study lexical distributions associated with FGD constructions to test existing claims in the literature.

\vspace{-5pt}
\paragraph{Generalization in computational models}
In addition to the filtered corpus training study presented in Section~\ref{sec:computational}, our detector enables input-attribution analyses, which aim to quantify which parts of the input or which training examples most causally affect a model’s prediction, typically by measuring output changes under perturbations or by computing gradient- or influence-based relevance scores (e.g., \citealp{koh2017understanding, hao-2020-evaluating, grosse2023studyinglargelanguagemodel}). Our detector enables attribution to aggregated groups of training data, linking model performance to construction-specific exposure. 

Taken together, these applications illustrate how fine-grained control and measurement of filler-gap input can help distinguish learning of shallow lexical patterns from learning of abstract dependency representations, and they provide a concrete bridge between acquisition-motivated corpus analysis and modern computational modeling.





\section*{Limitations}

Our detectors are imperfect due to the complexity of the data they take as input (e.g., there can be parser errors that mislead the detectors). Thus, the datasets we have extracted may not be suitable for research questions that require perfect identification of a given phenomenon. Nonetheless, we have shown that they are useful for illustrating broader statistical trends. 
Further, our detectors only identify some filler-gap phenomena and not others (e.g., cleftings and topicalization), and they are only defined over English. Future work could extend the approach to other phenomena and languages.

Our BabyLM filtered corpus training experiment only tested construction level corpus ablation. Future research could extend this approach by further analyzing the behavioral results and underlying mechanisms of controlled corpus filtering at different levels of granularity. One example is to filter the corpus by extraction site in order to test whether the FGD knowledge acquired for subject gaps can be generalized to object ones, and vice versa.
Future work can also experiment with alternative control dataset schemes such as removing only from the set of sentences that lack certain construction labels, as well as designing more fine-grained evaluations by testing subtypes of each construction. 

\section*{Acknowledgements}
We would like to thank the anonymous reviewers from CoNLL and SCiL 2026 for their thoughtful comments, which helped us refine this paper. We would also like to thank Athulya Aravind for helpful discussion and suggestions. We thank the Yale Center for Research Computing for guidance and use of the research computing infrastructure, specifically the Grace cluster. Any errors are our own.

\bibliography{parsing_childes}

\appendix

\section{An Example of How Dependency Parses Complement Constituency Parses}
\label{app:dep_cons_complementary}
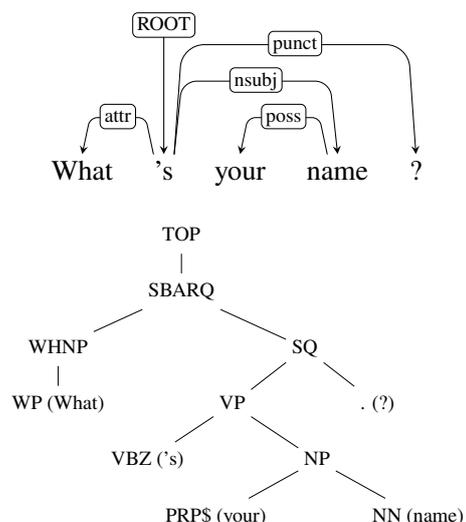
\begin{figure}[h]
  \centering
  \resizebox{0.7\columnwidth}{!}{\input{trees/dep_name.tex}}\par\medskip
  \resizebox{0.8\columnwidth}{!}{\input{trees/cons_name.tex}}
  \caption{Dependency parse (top) and constituency parse (bottom) of object matrix question \textit{What's your name?}}
  \label{fig:omq_demo}
\end{figure}

Section~\ref{subsec:hybriding} illustrates an example of how constituency parsing complements dependency parsing in distinguishing matrix questions from embedded questions. Here, we show an example of how the latter complements the former. For the object matrix question \textit{What's your name} (with an assumed deep structure of \textit{Your name is \underline{\hspace{1em}}}), as shown in Figure~\ref{fig:omq_demo}, the constituency parse has \texttt{SQ -> VP .} without an \texttt{NP} preceding \texttt{VP}, suggesting a subject gap. However, the dependency parse marks \textit{name} as the \texttt{nsubj}, suggesting the correct extraction site of object position. 
Similar examples exist for subtyping relative clauses.

There are ambiguities on how to interpret extraction sites. For instance, one may analyze the deep structure of Figure~\ref{fig:omq_demo} as \textit{\underline{\hspace{1em}} is your name}, so that this sentence is instead a subject matrix question. Such ambiguity is reflected in the inconsistent information given by the two parses. In these cases, we have to make a choice to implement the detector. Future studies can customize the heuristics to give different labeling if they choose different interpretations.

An additional practical motivation is imperfections in the source data: child speech can be noisy, ungrammatical, and subject to transcription errors, which could lead to parsing errors. Combining information from both parses makes the detection more robust to noise.

\section{An Elaboration of the Relative Clause Detection Algorithm}
\label{app:rc_detection}

\begin{figure*}[t]
  \centering
  \includegraphics[width=0.85\textwidth]{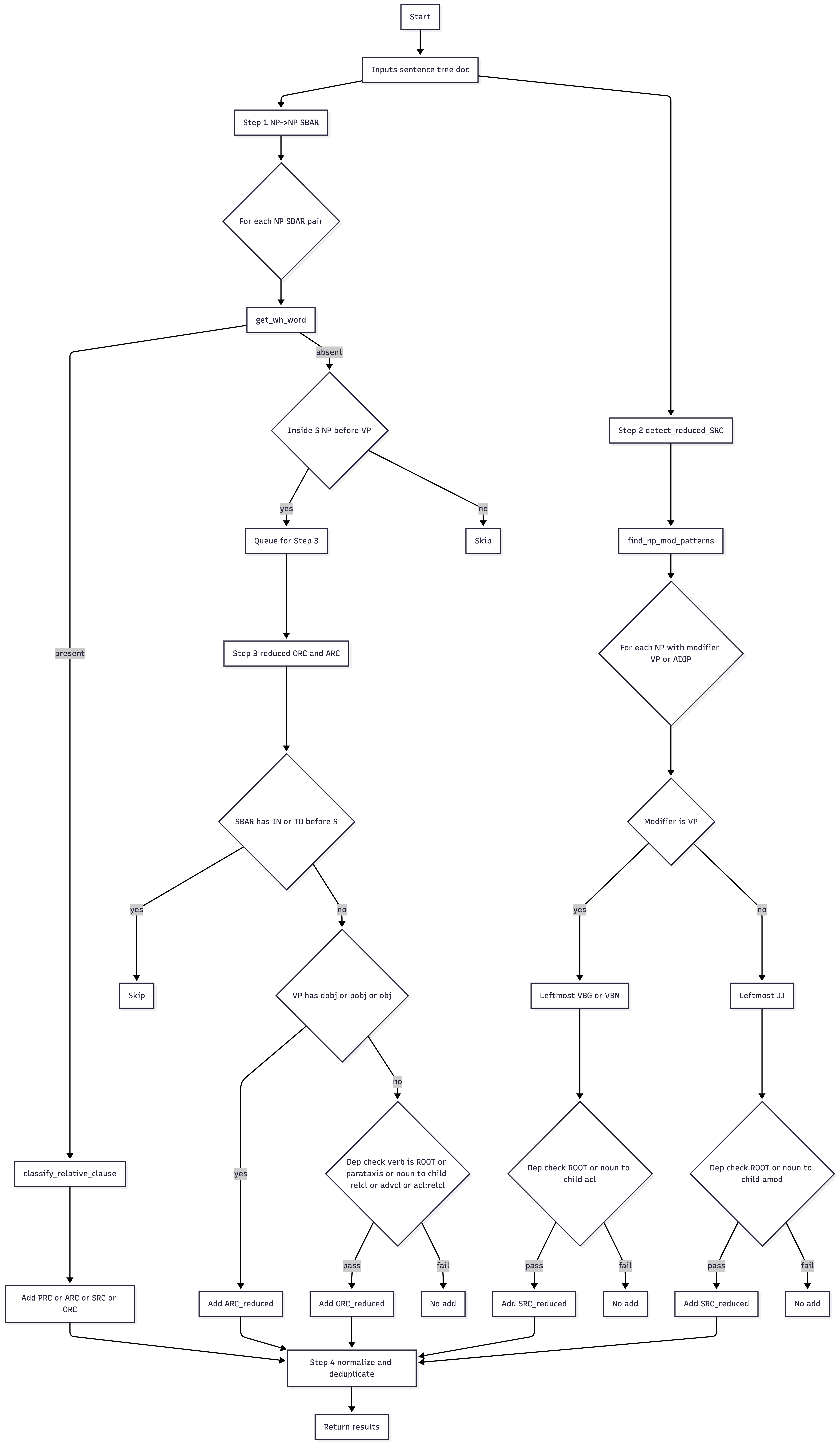}
  \caption{A flowchart for the relative clause detection process.}
  \label{fig:rc_diagram}
\end{figure*}

Figure~\ref{fig:rc_diagram} illustrates the decision flow for the types of relative clauses detected by the current detector.

\section{Additional Heuristics for Embedded Question Detection}
\label{app:embQ}
The distinction between embedded questions and free relatives depends on the selectional properties of the matrix predicate: question-embedding predicates select interrogative complementizer phrases (CP), while predicates that do not select questions force the wh-clause to be interpreted as a free relative (e.g.,~\citealp{huddleston2002cambridge}). For instance, \textit{I \textbf{know} [what he ate]} contains an embedded question since the matrix verb \textit{know} selects an interrogative CP, while the minimally different sentence \textit{I \textbf{ate} [what he ate]} contains a relative clause since \textit{eat} does not take an embedded question as its complement, so that this sentence is interpreted as \textit{I ate [the thing that he ate]}. This poses a challenge to our approach to identifying such structures using constituency and dependency parsing, since the resulting representations do not distinguish these two kinds of sentences.

In order to distinguish embedded questions from free relatives, we applied a lexical filter following the application of our constituency- and dependency-based heuristics for the detection of embeddeded questions. In practice, we applied constituency and dependency parsing to the entire BabyLM 10M dataset to identify all potential sentences involving an embedded-question-like structure. We then grouped them by the matrix verb lemma and sorted the verb lemmas by frequency. For each of the top 50 verb lemmas (covering 95\% of the total sentences with potential embedded question structure), we manually analyzed some example sentences containing that verb to decide if it can head embedded questions, resulting in the following list of verb lemmas:
\textit{know, see, tell, look, remember, wonder, guess, ask, say, forget, figure, understand, decide, show, watch, hear, think}. As the final step of our embedded question detection process, the detector only includes sentences with the selecting verbs in the list, and it additionally excludes sentences with \textit{whatever, whoever, whomever, whichever, whenever}, or \textit{wherever} in the embedded clause.
\section{Processing Treebank Annotations from \citeauthor{pearl2013computational}}
\label{app:ps13_details}

This appendix describes the process of comparing our detection results with Treebank annotations in \citet{pearl2013computational}. Figure~\ref{fig:ps13_tree_trace} shows a sample annotation for object matrix question \textit{what should the birdie say?}. Starting with manually corrected constituency parses from the Stanford parser\footnote{https://nlp.stanford.edu/software/lex-parser.shtml}, \citeauthor{pearl2013computational} marked gaps by annotating a trace item \texttt{*T*-1} with an index at the gap positions, together with the type of trace \texttt{-NONE-ABAR-WH-} (indicating a wh-movement), and they annotated the filler node \texttt{WHNP-1} to coindex with the trace. Relevant to the current study, there were 8157 wh-movement traces and 1,804 relative clause traces from their annotations.

\input{supp_tex_files/ps13_tree_example.tex}

The trace type and coindexing information enable us to determine extraction sites with high precision, which neither constituency nor dependency parsing explicitly provides. To directly compare annotated sentences with our detection results, we started with the Treebank-formatted annotations and developed simple heuristics to classify sentences into our labels given the parent nodes of the filler and the gap as well as their constituent types. We illustrate with the same example in Figure~\ref{fig:ps13_tree_trace}. We first checked the parent node of the gap, in this case \texttt{NP}, as well as the filler category, in this case \texttt{WHNP}, suggesting that the wh-word moves from an argument position. Next, we checked the first sentential complement node above the filler, in this case the \texttt{SBARQ} directly under \texttt{ROOT}, suggesting a matrix question. Finally, we traced down the sister node of the filler, in this case, \texttt{SQ}. Similar to Step 3 in Section~\ref{sec:detection_algo}, we checked whether there is an \texttt{NP} preceding the \texttt{VP} node, neglecting intermediate layers that contains auxiliary categories like \texttt{MD} (for modal, \textit{should}). In this case, the \texttt{NP} \textit{the birdie} precedes \texttt{VP} \textit{say}, this indicates that the subject position was already filled, suggesting an object gap.

\section{Processing CHILDES Data}
\label{app:childes_processing}

Given the objective of separating child versus adult speech, we extracted each speaker's age information from each transcript and corrected the speaker age of each utterance's metadata to reflect speech from children that were not labeled as \texttt{Target Child}.

Since the input distribution is likely to diverge between spontaneous production and targeted activities such as book reading, we further extracted metadata for each transcript via directly accessing raw files other than \texttt{.cha} files from the TalkBank.\footnote{https://talkbank.org/childes/access/Eng-NA/} This allowed us to do fine-grained disaggregations by longitudinal versus cross-sectional studies, as well as by daily, spontaneous productions versus various elicited productions.

See the released code base for the detailed transcription-level metadata. Future studies are encouraged to further explore more fine-grained results after disaggregation by types of studies and activities.

\section{Supplementary Data for CHILDES Statistics} \label{app:more_individual_results}

\begin{table}[h]
  \centering
  \small
  \setlength{\tabcolsep}{4pt}
  \renewcommand{\arraystretch}{1.05}
  \resizebox{\columnwidth}{!}{%
    \begin{tabular}{lcc}
      \toprule
      Pair & Adult $\Delta_{\mathrm{subj}}$ [95\% CI] & Child $\Delta_{\mathrm{subj}}$ [95\% CI] \\
      \midrule
      MatrixQ$\sim$EmbQ & [-0.057, 0.013] (n=25) & [-0.032, 0.102] (n=19) \\
      MatrixQ$\sim$RC & [-0.321, -0.256] (n=25) & [-0.309, -0.193] (n=20) \\
      EmbQ$\sim$RC & [-0.288, -0.239] (n=25) & [-0.322, -0.258] (n=19) \\
      \bottomrule
    \end{tabular}
  }%
  \caption{Across-age summary of $\Delta_{\mathrm{subj}}$ for each construction pair. Intervals are bootstrap 95\% CIs; parentheses give the number of age bins passing the minimum-count filter.}
  \label{tab:delta_subj_summary}
\end{table}

To complement the pairwise significance test of how object biases compare across constructions, we also summarize direct assessment of \textit{cross-construction} differences in Table~\ref{tab:delta_subj_summary}.
For each age bin and each pair of constructions, in addition to the subject share $p(\text{subj}) = \frac{\# \text{subject extractions}}{\# \text{subject extractions} + \# \text{object extractions}}$ described in Section~\ref{subsec:so_ratio}, we also took $\Delta_{\text{subj}}(c1,c2)=p_{c1}(\text{subj})-p_{c2}(\text{subj})$ (where $c$ stands for a construction), averaging across bins that meet a minimum-count threshold.
This cross-family summary shows that matrix and embedded questions are closely matched (mean $\Delta_{\text{subj}} \approx 0$), whereas relative clauses reliably differ from both question types, exhibiting a higher subject share ($\Delta_{\text{subj}} < 0$ for MatrixQ$\sim$RC and EmbQ$\sim$RC). Together, we conclude that extraction biases are not uniform across filler-gap constructions.


\section{Additional Details for BabyLM Filtered-Corpus Training} \label{app:babylm}
We conducted the filtered corpus training experiment outlined in Section~\ref{sec:computational} with two model architectures, GPT-2 and Llama. 
The relevant model configurations and training (hyper)parameters are summarized in Table~\ref{tab:training_config}.

\input{supp_tex_files/training_config_table.tex}

\begin{figure}[!h]
  \centering
  \includegraphics[width=0.95\columnwidth]{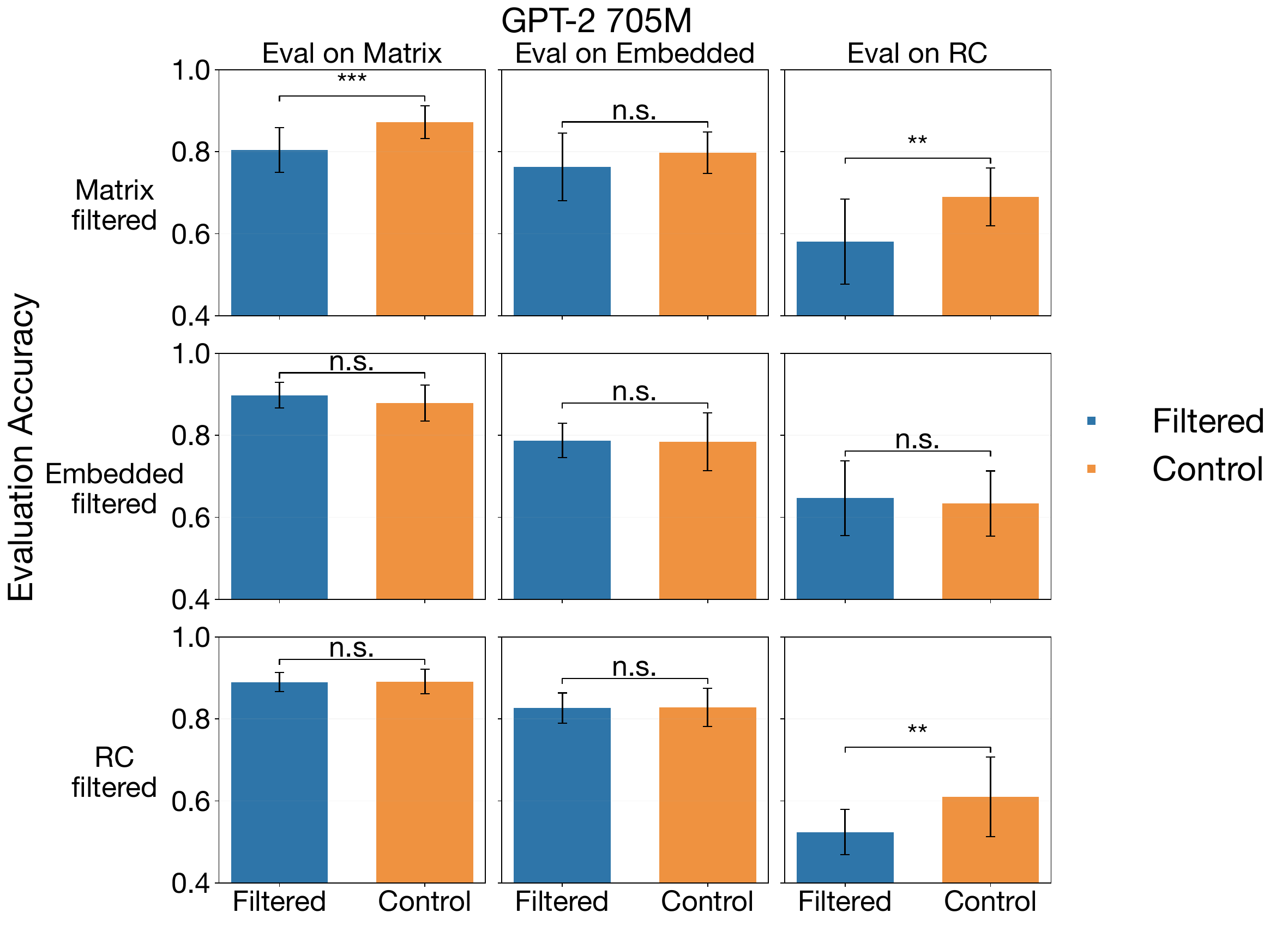}
  \caption{Cross-construction evaluations on filtered corpus training results of 15 GPT-2 models.}
  \label{fig:gpt2_results}
\end{figure}

In addition to the previously reported Llama figure, we also have results for GPT-2, which are shown in Figure~\ref{fig:gpt2_results}. Consistent with results for Llama (see Figure~\ref{fig:babylm}), we observe a significant decrease in performance when filtering with matrix questions and relative clauses and evaluated on the same construction. We nonetheless did not observe significant effects for embedded questions. One potential explanation is that matrix questions become the source of generalization as they have the highest frequency in the training data, compared to the other constructions. That is, it is easier to generalize from matrix questions to other constructions, but not in the other direction.

\section{Additional Details for the BabyLM Evaluation Dataset} \label{app:babylm_eval}

To construct the evaluation dataset, we created templates for each construction type.  Each template consists of three sentence components: two contexts and one continuation whose probability we test following each of the contexts.   In each case, the continuation is grammatical after context 1 but not after context 2. See our GitHub for all of the templates and the process for filling in the templates.

Note also that we did not include subject relative clauses, because there is ambiguity that is not caused by object relative clauses. For example in a minimal pair such as "I knew the author that wrote this book" and "I knew that wrote this book", the second sentence could be considered grammatical if we consider "that" as a demonstrative. To avoid this ambiguity, we did not include subject relative clauses in the evaluation dataset.
\end{document}

%% file: trees/dep_rc_demo.tex
\begin{dependency}
  \begin{deptext}[column sep=1em]
    I \& love \& the \& professor \& who \& taught \& syntax \& . \\
  \end{deptext}
  \depedge{2}{1}{nsubj}
  \depedge{4}{3}{det}
  \depedge{2}{4}{dobj}
  \depedge{6}{5}{nsubj}
  \depedge{4}{6}{relcl}
  \depedge{6}{7}{dobj}
  \depedge{2}{8}{punct}
  \deproot{2}{ROOT}
\end{dependency}

%% file: trees/cons_rc_demo.tex
\begin{forest}
    for tree={s sep=12mm, l sep=1mm, inner sep=1pt, align=center}
    [TOP [S [NP [PRP (I)]] [VP [VBP (love)] [NP [NP [DT (the)] [NN (professor)]] [SBAR [WHNP [WP (who)]] [S [VP [VBD (taught)] [NP [NN (syntax)]]]]]]] [. (.)]]]
\end{forest}

%% file: trees/dep_embed.tex
\begin{dependency}
  \begin{deptext}[column sep=1em]
    I \& remember \& which \& book \& Mary \& wrote \& . \\
  \end{deptext}
  \depedge{2}{1}{nsubj}
  \depedge{4}{3}{det}
  \depedge{6}{4}{dobj}
  \depedge{6}{5}{nsubj}
  \depedge{2}{6}{ccomp}
  \depedge{2}{7}{punct}
  \deproot{2}{ROOT}
\end{dependency}

%% file: trees/dep_matrix.tex
\begin{dependency}
  \begin{deptext}[column sep=1em]
    Which \& book \& do \& I \& remember \& Mary \& wrote \& ? \\
  \end{deptext}
  \depedge{2}{1}{det}
  \depedge{7}{2}{dobj}
  \depedge{5}{3}{aux}
  \depedge{5}{4}{nsubj}
  \depedge{7}{6}{nsubj}
  \depedge{5}{7}{ccomp}
  \depedge{5}{8}{punct}
  \deproot{5}{ROOT}
\end{dependency}

%% file: trees/cons_embed.tex
\begin{forest}
    for tree={s sep=9mm, l sep=1mm, inner sep=1pt, align=center}
    [TOP [S [NP [PRP (I)]] [VP [VBP (remember)] [SBAR [WHNP [WDT (which)] [NN (book)]] [S [NP [NNP (Mary)]] [VP [VBD (wrote)]]]] [. (.)]]]]
\end{forest}

%% file: trees/cons_matrix.tex
\begin{forest}
  for tree={s sep=3mm, l sep=2mm, inner sep=1pt, align=center}
  [TOP [SBARQ [WHNP [WDT (Which)] [NN (book)]] [SQ [VBP (do)] [NP [PRP (I)]] [VP [VB (remember)] [SBAR [S [NP [NNP (Mary)]] [VP [VBD (wrote)]]]]] [. (?)]]]]
\end{forest}

%% file: trees/dep_name.tex
\begin{dependency}
  \begin{deptext}[column sep=1em]
    What \& 's \& your \& name \& ? \\
  \end{deptext}
  \depedge{2}{1}{attr}
  \depedge{4}{3}{poss}
  \depedge{2}{4}{nsubj}
  \depedge{2}{5}{punct}
  \deproot{2}{ROOT}
\end{dependency}

%% file: trees/cons_name.tex
\begin{forest}
    for tree={s sep=18mm, l sep=1mm, inner sep=3pt, align=center}
    [TOP [SBARQ [WHNP [WP (What)]] [SQ [VP [VBZ ('s)] [NP [PRP\$ (your)] [NN (name)]]] [. (?)]]]]
\end{forest}

%% file: supp_tex_files/ps13_tree_example.tex
\begin{figure}[h]
  \small
  \ttfamily
  \setlength{\fboxsep}{6pt}
  \fbox{%
    \begin{minipage}{0.97\columnwidth}
\begin{tabbing}
(ROOT\\
\quad (SBARQ\\
\quad\quad \textbf{(WHNP-1-<INANIM>-<THEME-V1>}\\
\quad\quad\quad \textbf{(WP what))}\\
\quad\quad (SQ\\
\quad\quad\quad (VP\\
\quad\quad\quad\quad (MD should)\\
\quad\quad\quad\quad (NP-<ANIM>-<AGENT-V1>\\
\quad\quad\quad\quad\quad (DT the)\\
\quad\quad\quad\quad\quad (NN birdie))\\
\quad\quad\quad\quad (VP\\
\quad\quad\quad\quad\quad (VB-<V1> say)\\
\quad\quad\quad\quad\quad (NP\\
\quad\quad\quad\quad\quad\quad \textbf{(-NONE-ABAR-WH- *T*-1)))))}\\
\quad\quad (. ?)))\\
\end{tabbing}
    \end{minipage}%
  }
  \caption{Sample annotation with trace (bolded) from \citet{pearl2013computational}.}
  \label{fig:ps13_tree_trace}
\end{figure}

%% file: supp_tex_files/training_config_table.tex
\begin{table}[h]
\centering
\small
\setlength{\tabcolsep}{8pt}
\renewcommand{\arraystretch}{1.15}
\begin{tabular}{lcc}
\toprule
\textbf{(Hyper)parameter} & \textbf{GPT-2} & \textbf{Llama} \\
\midrule
\multicolumn{3}{l}{\textit{Data}} \\
Sequence length         & 128 & 128 \\
Eval samples            & 8192 & 8192 \\
\addlinespace

\multicolumn{3}{l}{\textit{Model}} \\
\# Parameters            & 705M & 360M \\
Hidden size             & 1536 & 1024 \\
Intermediate size       & 3072 & 3072 \\
\# Layers               & 24 & 24 \\
\# Heads                & 16 & 8 \\
Residual dropout        & 0.0 & -- \\
Attention dropout       & 0.0 & -- \\
Embedding dropout       & 0.0 & -- \\
\addlinespace

\multicolumn{3}{l}{\textit{Training}} \\
Learning rate           & $2.5\times 10^{-4}$ & $3.0\times 10^{-4}$ \\
Batch size              & 128 & 128 \\
\# Epochs               & 6 & 6 \\
Grad. accumulation      & 16 & 8 \\
Warmup steps            & 300 & 300 \\
FP16                    & True & True \\
\addlinespace

\bottomrule
\end{tabular}
\caption{Summary of configuration and training hyperparameters.}
\label{tab:training_config}
\end{table}